\documentclass{article}

\usepackage[preprint]{neurips_2020}

\usepackage[utf8]{inputenc} 
\usepackage[T1]{fontenc}    
\usepackage{hyperref}       
\usepackage{url}            
\usepackage{booktabs}       
\usepackage{amsfonts}       
\usepackage{nicefrac}       
\usepackage{microtype}      
\usepackage{graphicx}
\usepackage{amsmath}
\DeclareMathOperator*{\argmax}{\arg\!\max}

\usepackage{algorithm}
\usepackage{algpseudocode}
\usepackage{footnote}
\usepackage{tabularx}
\usepackage[utf8]{inputenc}
\usepackage[english]{babel}
\usepackage{csvsimple}

\title{TrialGraph: Machine Intelligence Enabled Insight from Graph Modelling of Clinical Trials}

\author{%
  Christopher Yacoumatos\\
  Data Science \& Artificial Intelligence, Biopharma R\&D, AstraZeneca\\
  Faculty of Mathematics, University of Cambridge\\
  Cambridge, UK\\
  \And
  Stefano Bragaglia\\
  Data Science \& Artificial Intelligence, Biopharma R\&D, AstraZeneca\\
  Cambridge, UK
  \And
  Anshul Kanakia\\
  Data Science \& Artificial Intelligence, Biopharma R\&D, AstraZeneca\\
  Gaithersburg, USA\\
  \And
  Nils Svangård\\
  Data Science \& Artificial Intelligence, Biopharma R\&D, AstraZeneca\\
  Gothenburg, Sweden\\
  \And
  Jonathan Mangion\\
  Data Science \& Artificial Intelligence, Biopharma R\&D, AstraZeneca\\
  Cambridge, UK\\
  \And
  Claire Donoghue\\
  Data Science \& Artificial Intelligence, Biopharma R\&D, AstraZeneca\\
  Cambridge, UK\\
  \And
  Jim Weatherall\\
  Data Science \& Artificial Intelligence, R\&D, AstraZeneca\\
  Macclesfield, UK
  \And
  Faisal M. Khan\\
  Data Science \& Artificial Intelligence, Biopharma R\&D, AstraZeneca\\
  Gaithersburg, USA
  \And
  Khader Shameer\thanks{Corresponding Author: shameer.khader@astrazeneca.com}\\
  Data Science \& Artificial Intelligence, Biopharma R\&D, AstraZeneca\\
  Gaithersburg, USA}

\begin{document}

\maketitle

\newpage

\begin{abstract}
  A major impediment to successful drug development is the complexity, cost, and scale of clinical trials. The detailed internal structure of clinical trial data can make conventional optimization difficult to achieve. Recent advances in machine learning, specifically graph-structured data analysis, have the potential to enable significant progress in improving clinical trial design. TrialGraph seeks to apply these methodologies to produce a proof-of-concept framework for developing models which can aid drug development and benefit patients. In this work, we first introduce a curated clinical trial data set compiled from the CT.gov, AACT and TrialTrove databases (n=1191 trials; representing one million patients) and describe the conversion of this data to graph-structured formats. We then detail the mathematical basis and implementation of a selection of graph machine learning algorithms, which typically use standard machine classifiers on graph data embedded in a low-dimensional feature space. We trained these models to predict side effect information for a clinical trial given information on disease, existing medical conditions, and treatment. The MetaPath2Vec algorithm performed exceptionally well, with standard Logistic Regression, Decision Tree, Random Forest, Support Vector, and Neural Network classifiers exhibiting typical ROC-AUC scores of 0.85, 0.68, 0.86, 0.80, and 0.77, respectively. Remarkably, the best performing classifiers could only produce typical ROC-AUC scores of 0.70 when trained on equivalent array-structured data. Our work demonstrates that graph modelling can significantly improve prediction accuracy on appropriate datasets. Successive versions of the project that refine modelling assumptions and incorporate more data types can produce excellent predictors with real-world applications in drug development.
\end{abstract}

\section{Introduction}\label{sec1}

The process of developing new drugs is arduous, time-consuming and expensive. It typically costs in excess of \$1 billion to bring a drug from the early discovery stage through clinical trials to regulatory approval and mass production, and the whole process is rarely completed within a decade. Furthermore, only 5\% of initial drug programs successfully complete this process [1]; the vast majority of candidate drugs produce no revenue and offset the profits of successful drugs. As a result, pharmaceutical companies are keen to develop methods to reduce the operating costs and timescales involved in order to produce drugs more cheaply and rapidly, which ultimately stands to benefit the patients who need them [2]. There is particular interest in streamlining the clinical trial phase of the drug development process. Clinical trials are becoming increasingly complex due to shifts in regulatory requirements, and a review of oncology trials from 1996 to 2016 showed that there were significantly more pharmacokinetic requirements to be fulfilled in 2016 than in 1996 [3]. Despite the fact that large amounts of high-quality data are collected as part of the clinical trial process, trial sponsors typically struggle to use this data to improve clinical trial design and reduce trial cost and scale. However, recent advances in machine learning, specifically in the field of graph structured data analysis, have enabled significant progress.

The term 'machine learning' is a broad term which describes a range of predictive algorithms whose internal hyperparameters are iteratively updated to minimise the prediction error. As many problems in medicine can be rephrased in terms of a prediction problem, applications of machine learning are ubiquitous in the field. A standard example is that of skin cancer detection, where machine learning algorithms are deployed to produce a model which uses information about the physical characteristics of skin lesions to classify them into benign and malignant classes. However, traditional machine learning techniques can fail when applied to certain complex data structures. Datasets which demonstrate a high level of connectivity between individual data points are typically best represented by networks, such as biomedical target-pathway networks, for which graph-based machine learning algorithms are required to interpret the data and potentially improve prediction analytics [4-6].

It is possible to represent clinical trial data in both traditional array and graph-structured data formats. Since information about each separate clinical trial is recorded, clinical trial data can naturally be represented by an array whose rows represent each clinical trial and whose columns represent data types. However, since some data types are discrete e.g. medical condition information, clinical trials can be considered connected if they share the same discrete data value e.g. patients share the same medical conditions. The methodology behind graph modelling clinical trial data is developed later in the Data section. The development of machine learning techniques which produce good results for complex data such as clinical trial data is important for streamlining the clinical trial process [7]. For example, it can allow us to develop models that can predict which pre-clinical drugs are most likely to fail clinical trials. Pharmaceutical companies can then focus their efforts on the drugs that are most likely to succeed, thereby avoiding the extra cost and delay of a failed clinical trial.

\subsection{Manuscript Scope \& Aim}

In this manuscript, we document the progress of the TrialGraph project, which has multiple aims.
Primarily, TrialGraph seeks to successfully use graph-based machine learning techniques on a curated clinical trial data set (n=1191 trials; representing one million patients) to produce a model which can predict side effect information given information on disease, existing medical conditions and treatment.
It then aims to compare the performance of the most successful graph-based machine learning techniques to that of equivalent techniques on equivalent data in order to evaluate whether graph modelling can boost the predictive power of machine learning techniques on data with a similar structure to clinical trial data.
TrialGraph also seeks to serve as a brief review of graph-based machine learning techniques.

The manuscript is structured as follows: in Section 2 we introduce the curated clinical trial data set, and describe the methodologies behind graph modelling and the process of converting the data into a graph format. Section 3 describes the mathematical background of a selection of graph machine learning algorithms and their implementation on the graph-structured clinical trial data. In Section 4, we provide the results observed when the algorithms were used to predict side effect information for test data, with a comparison of the performance of equivalent algorithms on equivalent array-structured data. Section 5 evaluates the overall success of graph-based machine learning algorithms and discusses limitations, as well as applications in industry and areas of further exploration.

\section{Data}

In this section, we introduce the curated clinical trial dataset used in TrialGraph and describe its key features. We highlight the connections the dataset draws between separate clinical trials and provide a methodology for converting this dataset into a network, along with schemas for the networks to which we apply graph-based machine learning methods.

\subsection{Clinical Trial Dataset}

The dataset we use describes clinical trial data for 1191 clinical trials, representing one million patients. Data was obtained from the AACT, CT.gov and TrialTrove databases, and was cleaned and processed prior to use. Table \ref{table1} provides a summary of the categories of data recorded for each clinical trial, with a brief description for each category. The potential for connections between clinical trials is apparent; if participants in two separate clinical trials suffer from Pulmonary Fibrosis and Pulmonary Hypertension respectively, then the clinical trials ought to be linked by the fact that participants from both trials suffer from pulmonary diseases. To facilitate the formation of connections between clinical trials, we formalise this approach. For all Disease and Existing Medical Conditions data entries, we extract keywords i.e. phrases that appear within multiple entries. We then form the 'Condition' set as the set of all such keywords, and the 'Specific Condition' set as the set of terms that appear under Disease and Existing Medical Conditions (excluding keywords). We apply the same procedure to Primary Tested Drug, to obtain the 'Drug' and 'Specific Drug' sets.

\begin{table}
  \caption{Data Category Dictionary}
  \label{table1}
  \centering
  \begin{tabular}{lll}
    \toprule
    Category & Sample Value & Description \\
    \midrule
    NCT\_id & NCT00001596 & Unique Identifier assigned to a trial by clinicaltrials.gov \\
    Trial ID & 161979 & Unique Identifier assigned to a trial by trialtrove.gov \\
    Disease & Pulmonary & Disease for which patients in a clinical trial were treated \\
    AE\_AST\_ALT & 0.028571429 & Percentage presenting high AST and/or ALT\\
    AE\_COPD & 0.575600801 & Percentage presenting COPD\\
    ... & ... & ... (62 total Adverse Events) ...\\
    AE\_Thrombosis & 0.041666667 & Percentage presenting thrombosis\\
    MeSH Term & Sleep Apnea & Existing medical condition for patients in a clinical trial\\
    Drug & Pirfenidone & Drug used to treat patients in a clinical trial\\
    \bottomrule
  \end{tabular}
\end{table}

Before we demonstrate examples of graph-structured clinical trial data, we first define graph-related terminology and concepts.

\newpage

\subsection{Graph Theory}

Definition (Graph): A graph $G$ is a tuple $G=(V,E)$ where $V$ is a set of $\|V\|$ many nodes and $E\subset V\times V$ is a set of edges linking two nodes in $V$. A graph is directed if $(a,b)\in E$ denotes an edge with direction from node $a$ to node $b$, otherwise the graph is undirected, and edges do not have assigned directions. While the graphs considered in this work have edges which have a contextual direction, we do not investigate directionality in this analysis, and treat our graphs as undirected.

Definition (Attributed graph): An attributed graph is a graph where labels and/or attributes are assigned to nodes and/or edges. Where present, labels are assumed to be defined over a finite set, $\Sigma_{V}$ for nodes and $\Sigma_{E}$ for edges, with node and edge labelling functions
\[\mathbb{L}_{V}:V\rightarrow \Sigma_{V}\qquad
\mathbb{L}_{E}:E\rightarrow \Sigma_{E}\]
where both functions are total and surjective.

Similarly, attributes are defined over $\textrm{R}^{d}$ for some $d$, with node and edge attributing functions 
\[\mathbb{A}_{V}:V\rightarrow \textrm{R}^{d}\qquad
\mathbb{A}_{E}:E\rightarrow \textrm{R}^{d}\]
where both functions are total.

Note that from these definitions, the graph representation of a dataset is not unique. A graph can be reconfigured to incorporate label information as attribute information and vice versa; indeed each graph machine learning method applied in TrialGraph requires clinical trial data to be configured in different ways. We now describe the schemas for each of these configurations.

\subsection{Complete Knowledge Graph}

This large undirected graph is node and edge labelled, but node and edge unattributed. It represents information on disease, existing medical conditions, side effects and drugs for all available clinical trials. Table \ref{table2} describes node and edge labels and connection criteria, and Figure 1 provides a graph schematic.

\begin{table}
  \caption{Complete Knowledge Graph Summary}
  \label{table2}
  \centering
  \begin{tabularx}{\textwidth}{|c|X|c|}
    \toprule
    Node Label & Description & Frequency\\
    \midrule
    \textbf{Clinical Trial} & - & 1178\\
    \textbf{Adverse Event} & - & 62\\
    \textbf{Drug} & Element of 'Drug' set & 109\\
    \textbf{Specific Drug} & Element of 'Specific Drug' set & 403 \\
    \textbf{Condition} & Element of 'Condition' set & 57 \\
    \textbf{Specific Condition} & Element of 'Specific Condition' set & 285 \\
    \midrule
    Edge Label & Connection Criteria & Frequency\\
    \midrule
    Expresses & Non-zero dropout percentage due to \textbf{Adverse Event} for a \textbf{Clinical Trial} & 22018 \\
    & & \\
    Diagnosis & \textbf{Condition} appears in \textbf{Clinical Trial} information & 4395 \\
    & & \\
    Specific Diagnosis & \textbf{Specific Condition} appears in \textbf{Clinical Trial} information & 4926 \\
    & & \\
    Treatment & \textbf{Drug} appears in \textbf{Clinical Trial} information & 854 \\
    & & \\
    Specific Treatment & \textbf{Specific Drug} appears in \textbf{Clinical Trial} information & 679 \\
    & & \\
    Condition Specification & \textbf{Condition} appears in \textbf{Specific Condition} & 57 \\
    & & \\
    Drug Specification & \textbf{Drug} appears in \textbf{Specific Drug} & 109 \\
    & & \\
    Treatment Targeting & \textbf{Specific Condition} in Disease for a clinical trial using \textbf{Drug} & 8347 \\
    \bottomrule
  \end{tabularx}
\end{table}

\begin{figure}[H]
  \centering
  \includegraphics[width=0.75\textwidth]{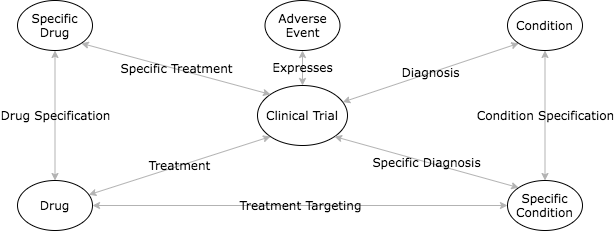}
  \caption{Complete Knowledge Graph Schema.}
\end{figure}

\subsection{Complete Bi-Nodal Graph}

This large undirected graph is node labelled and attributed. Edges are unlabelled but attributed. It represents information on disease, existing medical conditions, side effects and drugs for all available clinical trials. Table \ref{table3} describes node/edge labels and connection criteria, and Figure 2 provides a graph schematic.

\begin{figure}[H]
  \centering
  \includegraphics[width=0.75\textwidth]{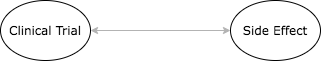}
  \caption{Complete Bi-Nodal Graph Schema.}
\end{figure}

\begin{table}
  \caption{Complete Bi-Nodal Graph Summary}
  \label{table3}
  \centering
  \begin{tabularx}{\textwidth}{|X|X|c|}
    \toprule
    Node Label & Attribute Description & Frequency \\
    \midrule
    \textbf{Clinical Trial} & For each \textbf{Clinical Trial}, information about Condition, Specific Condition, Drug and Specific Drug is stored as a large vector using one-hot encoding & 1178\\
    & & \\
    \textbf{Adverse Event} & For each \textbf{Adverse Event}, information about prevalence within the dataset is generated and stored as a vector & 62\\
    \midrule
    Edge Connection Criteria & Attribute Description & Frequency\\
    \midrule
    All pairs of \textbf{Clinical Trial} and \textbf{Adverse Event} are connected & Edges are attributed a weight of 1 if there is a non-zero dropout percentage due to \textbf{Adverse Event} for a \textbf{Clinical Trial}, 0 otherwise & 73036\\ 
    \bottomrule
  \end{tabularx}
\end{table}

\subsection{Constituent Knowledge Graph}

This is a set of graphs where each graph represents a clinical trial, and is node and edge labelled but node and edge unattributed. It shares the same node/edge labels and connection criteria as the Complete Knowledge Graph, and draws from the same collection of nodes and edges, but only features those relevant to the particular clinical trial.

\subsection{Network Representations of Graphs}

Network representations for the Complete Knowledge and Complete Bi-Nodal Graphs were generated using Cytoscape and are provided in Figure 3. For the Complete Knowledge Graph, edges are coloured by edge label, while for the Complete Bi-Nodal Graph, edges are included if they have a weight of 1.

\begin{figure}[H]
  \centering
  \includegraphics[width=\textwidth]{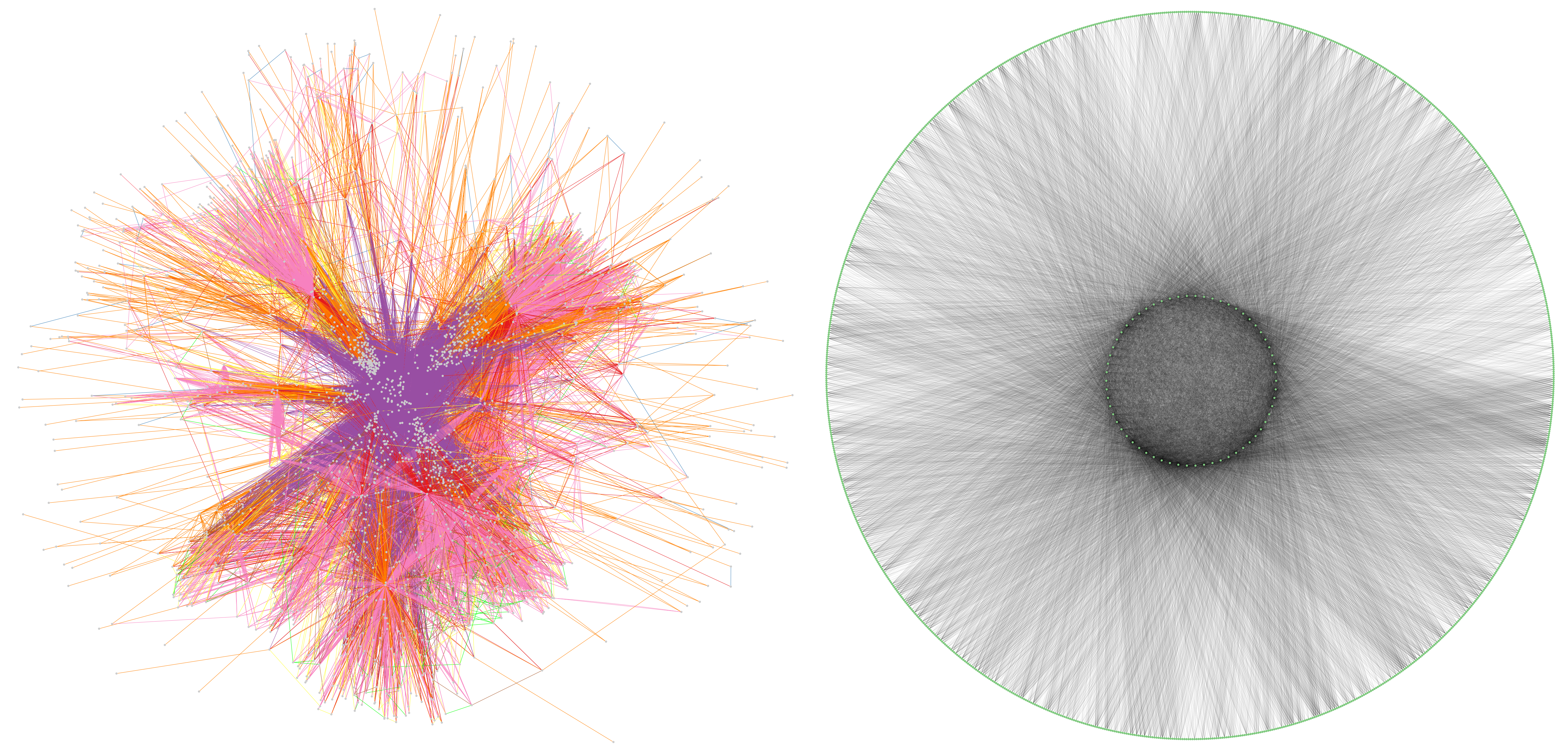}
  \caption{Network visualisations of the Complete Knowledge Graph and the Complete Bi-Nodal Graph respectively.}
\end{figure}

\section{Method}\label{sec3}

Most graph-based machine learning algorithms operate by a two-stage method. Typically, a general purpose encoder is first used to embed node and/or edge data into a low-dimensional feature space in such a way that preserves the original graph structure i.e. if two nodes in the graph can be considered 'close', then their representations in feature space should be 'close'. Then a task-specific decoder can be trained on these representations to produce predictions for node/edge classification/regression problems [8][9]. Under this framework, the encoding stage is of particular interest. Many different notions of 'closeness' can be defined, each with a corresponding encoder suitable for certain tasks.

The most basic encoders directly apply standard graph metrics to nodes and/or edges to produce low-dimensional representations. An example of such a metric is the k-nearest neighbours approach; for each node, it finds the number of nodes which can be reached by traversing at most k edges. If we repeat this for $1\leq k \leq n$, we can embed each node into an $n$-dimensional feature space. More refined approaches consider subgraph isomorphism between neighbourhoods of nodes, but these approaches can become computationally expensive as graph isomorphism tests are difficult to perform. Random walk approaches are a useful alternative. These work by considering a set of random walks along graph edges starting at a specific node. A natural measure of 'closeness' is given by the proportion of random walks which end up at a certain target node, and this can be used to generate a suitable embedding function. Graph Neural Networks operate slightly differently; these typically stack layers of nodes on top of the graph itself, with each layer determined by an aggregation of nodes from the previous layer. A collection of weights determining the aggregation functions can then be tuned to create a model which produces predictions.

Standard implementations of these approaches are only suitable for unlabelled graphs, as these methods are do not differentiate between nodes of different types. As our clinical trial dataset is inherently labelled, these methods must be modified before they can be applied to produce predictions. In this section, we introduce the MetaPath2Vec and HinSage methods, which are essentially versions of the Random Walk and Graph Neural Network approaches that have been modified to work on labelled graphs. We also introduce the Graph Kernel method, a leading-edge method which operates in a different manner to most existing methods.
The implementation details of these methods on the clinical trial dataset are also provided. Finally, we also include a standard machine learning approach on the same clinical trial data converted to an array format through one-hot encoding.

In order to provide a basis for comparison, all methods described above perform the same edge classification task: to predict whether a Clinical Trial has a non-zero dropout percentage due to an Adverse Event. This is a useful prediction task in the context of streamlining drug development, as if it were possible to determine which drugs would cause serious side effects in patients with a given diagnosis and medical history, then these side effects can be avoided to improve patient safety, and drugs can be removed from the development process without the need for clinical trials. While more detailed predictions could be generated through the use of regression, not all of the above methods can perform regression tasks, so classification tasks are used for comparison.

\subsection{MetaPath2Vec}

This method is an extension of the Skip-gram architecture, which seeks to find the objective function that maximizes the log-probability of finding a neighbourhood $N_{s}(u)$ for a node $u$ conditioned on the node embedding in the feature space [10]. For a graph $G=(V,E)$ this function $f$ is given by:

\[\argmax_{f}\sum_{u\in V} \textrm{log}(\mathbb{P}(N_{s}(u)\|f(u))\]

This problem is made tractable by two assumptions:

Conditional Independence. The likelihood can be factorized by assuming the probability of observing a neighbourhood node is independent of observing any other neighbourhood node given the embedding of the source node i.e. 

\[\mathbb{P}(N_{s}(u)\|f(u)) = \prod_{n_{i}\in N_{s}(u)}\mathbb{P}(n_{i}\|f(u))\]

Symmetry in feature space. A source node and neighbourhood node have a symmetric effect over each other in feature space. Hence we can model the conditional probability of every source-neighbourhood node pair as a softmax unit parametrised by a dot product of their features i.e. 

\[\mathbb{P}(n_{i}\|f(u))=\frac{\textrm{exp}(f(n_{i})\cdot f(u))}{\sum_{v\in V}\textrm{exp}(f(v)\cdot f(u))}\]

With these two assumptions, the problem is well posed under a suitable sampling strategy for $N_{s}(u)$, the neighbourhood of a node $u$. Two extremes exist for sampling neighbourhoods $N_{s}$ of $k$ nodes:
Breadth First Sampling restricts the neighbourhood of $u$ to the set of nodes connected to $u$, and samples $k$ nodes randomly from this set.
Depth First Sampling restricts the neighbourhood of $u$ to a sequential set of nodes of increasing distance from $u$, and samples the first $k$ nodes.

Random Walk Sampling lies between these extremes, and samples the nodes reached by a random walk between nodes connected by edges, of length $k$. The edge probabilities are determined inductively; if a random walk has progressed from nodes $u$ to $v$, then the (un-normalised) probability the walk progresses from $v$ to some node $w$ takes one of three user-defined values depending on whether $w$ is the same node as $u$, or if it is one or two edges away from $v$.

Once the optimal node embedding function $f$ has been obtained, an edge embedding function
\[g :V\times V \rightarrow \mathbb{R}^{d}\]
can be obtained as
\[g((a,b))=f(a)\odot f(b)\]
where $\odot$ is a suitable binary operator such as the Hadamard operator.

This approach is sufficient for homogeneous graphs but not for heterogeneous graphs, such as our clinical trial data set. If the random walk method is applied as described above, it can consider two graph nodes with different labels to be 'close' even though they are contextually unrelated. This then causes the optimal mapping function to incorrectly embed the nodes close to each other in the feature space. Drawing an explicit example from the Complete Knowledge Graph, the standard random walk method can consider nodes 'Pulmonary' and 'AE\_AST\_ALT' to be 'close' as they are connected by a path of length two via 'NCT00001596', even though these nodes refer to diseases and side effects and are unrelated in the context of the clinical trial data.

The MetaPath2Vec algorithm addresses this issue by allowing the user to define suitable 'metapaths' which predefine which edges the random walk should traverse in the context of the data.

\subsubsection{MetaPath2Vec Implementation}

The MetaPath2Vec algorithm was implemented on the Complete Knowledge Graph for the clinical trial dataset to perform edge prediction for edges of type 'Expresses' using StellarGraph [11]. From the original graph, 10\% of 'Expresses' edges were removed and combined with an equal number of 'negative' edges (i.e. pairs of Clinical Trial, Side Effect nodes which were not connected by an 'Expresses' edge) to form a test set of edges. A further 40\% of edges were removed to form a training set. The MetaPath2Vec algorithm with random walks of length 200 and a given set of metapaths\footnote{See Supplemental Data - (1)} was then run on the remaining graph to determine a good edge embedding function. This embedding function was applied to the edges in the training set to produce vector embeddings of dimension 512, and several standard machine learning classifiers (Logistic Regression, Random Trees and Random Forests, Support Vector and Neural Network) were trained on this data to classify edges as 'positive' or 'negative'.

The embedding function and trained classifiers were then applied to the test set, and ROC-AUC scores were generated for the classifiers' predictions and are provided in the Results section. A workflow diagram for the implementation of the MetaPath2Vec algorithm is given in Figure 4.

\begin{figure}
  \centering
  \includegraphics[width=\textwidth]{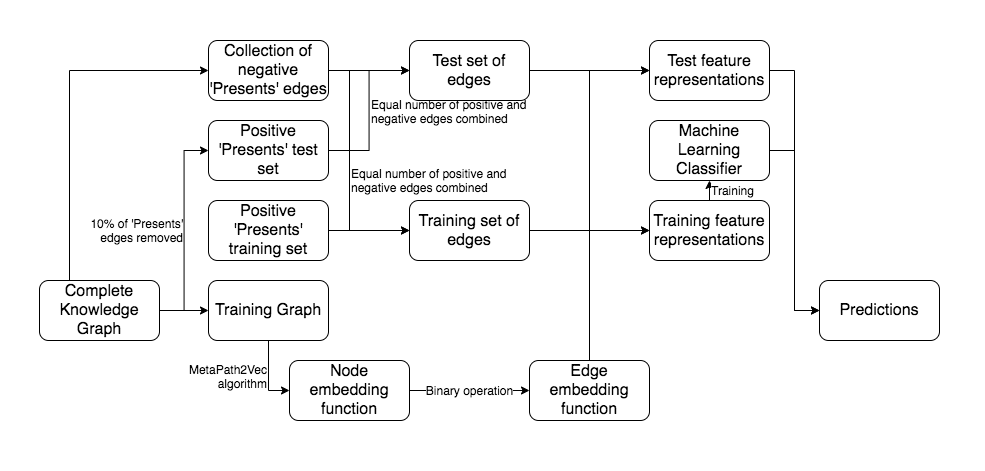}
  \caption{MetaPath2Vec workflow diagram.}
\end{figure}

\subsection{HinSAGE}

HinSAGE is an extension of the GraphSAGE algorithm which uses a standard GCN architecture with $k$ hidden layers to generate node embeddings. Intuitively, the algorithm aggregates information from increasingly distant neighbours as its search depth increases, and is given explicitly in algorithm 1 [12].

\begin{algorithm}
\caption{GraphSAGE embedding generation}\label{algo1}
\begin{algorithmic}[1]
\Require Graph $G=(V,E)$, node attributes $x_{v}$ for all $v \in V$, weight matrices $W_{1\leq i\leq k}$, non-linear activation $\sigma$, aggregator functions $f_{1\leq i\leq k}$ , neighbourhood function $N:v\rightarrow 2^V$
\Ensure Vector representations $z_{v}$ for all $v \in V$
\State $h_{v}^{0}=x_{v}$
\For{$i=1...k$}
    \For{$v\in V$}
        \State $h_{N(v)}^{i} \leftarrow f_{i}(h_{u}^{i-1},\forall u\in N(v)$
        \State $h_{v}^{i} \leftarrow \sigma(W_{i}\cdot\textrm{CONCAT}(h_{v}^{i-1},h_{N(v)}^{i}))$
    \EndFor
    \State $h_{v}^{i} \leftarrow \hat{h_{v}^{i}},\forall v\in V$
\EndFor
\State $z_{v} \leftarrow h_{v}^{k},\forall v\in V$
\end{algorithmic}
\end{algorithm}

The aggregator function used is the mean aggregator

\[h_{N(v)}^{i}=\frac{1}{\|N(v)\|}D_{p}[h_{u}^{i-1},\forall u\in N(v)]\]

where $D_{p}[\cdot]$ is a random dropout with probability $p$ applied to its argument vector.

The parameters of the aggregator functions and the weight matrices are tuned by enforcing nearby nodes to have similar feature representations, which is done by minimizing the objective function

\[\eta (z_{u}) = -\textrm{log}(\sigma(z_{u}^{\textrm{T}}z_{v}))-Q\cdot \mathbb{E}_{v_{n}\sim P_{n}(v)}\textrm{log}(\sigma(z_{u}^{\textrm{T}}z_{v}))\]

where $v$ is a node reachable from $u$ on a fixed-length random walk, $\sigma$ is the sigmoid function, $P_{n}$ is a negative sampling distribution and $Q$ defines the number of negative samples. Once node embeddings are generated, edge embeddings can be determined using binary operators as in the MetaPath2Vec algorithm.

The standard GraphSAGE algorithm is only applicable for graphs which are unlabelled and fully node attributed (i.e. every node has an attribute). HinSAGE modifies this algorithm to make it compatible for node labelled and attributed, edge unlabelled and unattributed graphs; it defines new weight matrices for each pair of nodes linked by an edge and modifies the aggregator functions to make use of these weight matrices.

\subsubsection{HinSAGE Implementation}

The HinSAGE Algorithm was implemented on the Complete Bi-Nodal Graph for the clinical trial dataset to perform edge prediction using StellarGraph [11]. From the original graph, 10\% of edges were removed to form a test set. A standard GCN architecture was then built on the remaining graph, with two hidden layers of 128 nodes and a link classification layer on top. Internal parameters were optimized over 20 epochs using the ADAM optimizer. The classifier was then applied to the test set; ROC-AUC scores were generated for the classifier's predictions and are provided in the Results section. A workflow diagram for the implementation of the HinSAGE algorithm is given in Figure 5.

\begin{figure}
  \centering
  \includegraphics[width=\textwidth]{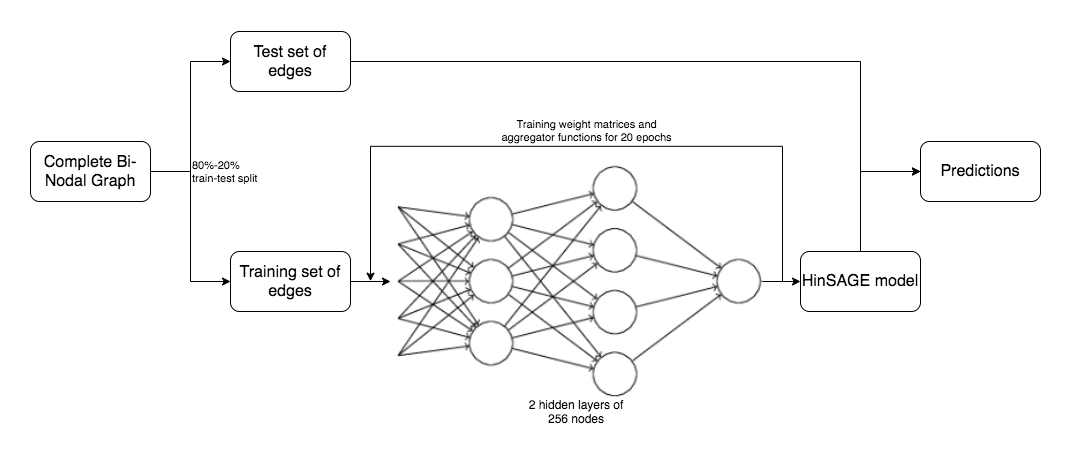}
  \caption{HinSAGE workflow diagram.}
\end{figure}

\subsection{Graph Kernel}

Graph Kernel methods are motivated by the limitations of subgraph isomorphism tests on neighbourhoods of nodes for use in generating graph metrics. It is known that subgraph isomorphism problems are NP-complete i.e. they cannot be solved in polynomial time, which means that their practical use is limited for applications which involve large quantities of data, such as machine learning on clinical trial data. Graph kernels aim to provide a measure of similarity between graphs in polynomial time, which can then be used to generate metrics for use in machine learning [13].

Definition (Kernel): Given a non-empty set $X$, a function $k: X\times X\rightarrow\mathbb{R}$ is a kernel if there exists a Hilbert space $\mathbb{H}$ and some map $\phi :X\rightarrow \mathbb{H}$ such that

\[k(x,y)=\langle \phi (x),\phi(y)\rangle_{\mathbb{H}}\quad\textrm{where}\quad\langle\cdot\rangle_{\mathbb{H}}\quad\textrm{is an inner product in}\quad \mathbb{H}\]

for all $x,y\in\mathbb{H}$. 

Definition (Symmetric positive definite bivariate function): Let $X$ be a set and $f:X\times X\rightarrow \mathbb{R}$ be a bivariate function. Then $f$ is symmetric if it satisfies
\[f(x,y)=f(y,x)\quad\forall x,y\in X\]
and $f$ is positive definite if it satisfies
\[\sum_{i=1}^{k}\sum_{j=1}^{k}\lambda_{i}\lambda_{j}f(x_{i},x_{j})\geq 0\quad\forall(\lambda_{1},...,\lambda_{k})\in\mathbb{R}^{k},\quad\forall(x_{1},...,x_{k})\in X^{k}\]

A corollary of the Moore-Aronszajn theorem states that a function $k: X\times X\rightarrow\mathbb{R}$ is a valid kernel if and only if it is symmetric and positive definite.

For $X=\mathbb{R}^{n}$, a standard example of a kernel function is the radial basis function kernel, defined as 
\[k_{\textrm{RBF}}(x,y)=\textrm{exp}(-\frac{\|x-y\|^{2}}{2\sigma^{2}})\]
for some scalar parameter $\sigma$. This can be used to define a very simple graph kernel. Let $\mathbb{G}$ be a set of graphs with node attributes, and let $G=(V_{G},E_{G})$, $H=(V_{H},E_{H})$ be elements of $\mathbb{G}$. Then the all node-pairs kernel $k_{N}:\mathbb{G}\times\mathbb{G}\rightarrow \mathbb{R}$ is defined as
\[k_{N}(G,H)=\sum_{u\in V_{G}}\sum_{v\in V_{H}}k_{\textrm{RBF}}(u,v)\]
After appropriate scaling, the range of the kernel is the interval $[0,1]$, where a value close to $0$ indicates graphs $G$ and $H$ are highly dissimilar and a value close to $1$ indicates graphs $G$ and $H$ are highly similar.
Many other graph kernels can be defined, and those that can distinguish node/edge labelling, such as the ODD-STh kernel, are particularly useful for the purposes of TrialGraph.

\subsubsection{Graph Kernel Implementation}

From the set of Constituent Knowledge Graphs, 50 graphs were removed to form a reference set. Training and test sets were formed from the remaining set of graphs using an 80\%-20\% split; several kernels suitable for node and edge labelled graphs (ODD-STh, SubgraphMatching, Propagation) were evaluated for each pair of graphs in the training and reference sets respectively using GraKel [14]. The values were stored in arrays for each kernel, along with information on whether each clinical trial associated with a training graph demonstrated a non-zero dropout percentage for a predetermined side-effect. Support vector classifiers were then trained on this data. The same kernels were evaluated for each pair of graphs in the test and reference sets, and the classifiers were applied to this data. ROC-AUC scores were generated for the classifiers' predictions under several kernels and are provided in the Results section. A workflow diagram for the implementation of the Graph Kernel algorithm is given in Figure 6.

\begin{figure}
  \centering
  \includegraphics[width=\textwidth]{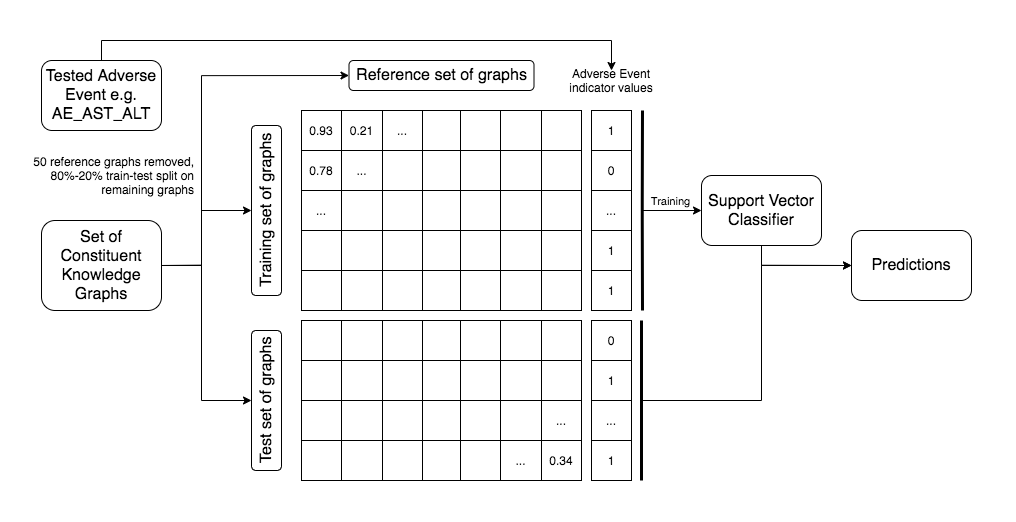}
  \caption{Graph Kernel workflow diagram.}
\end{figure}

\subsection{Array-Sturctured Comparison}

One of TrialGraph's aims is to determine whether structuring clinical trial data as a graph and applying graph-based machine learning techniques provides greater prediction power than implementing standard array-structured machine learning methods. To determine a fair comparison on array-structured data, we created an array where for each Clinical Trial, Adverse Event pair, we store Disease, Existing Medical Condition and Drug information associated with each Clinical Trial through one-hot encoding, and information on the Adverse Event prevalence within the whole dataset. Furthermore, for each Clinical Trial, Adverse Event edge, we record an indicator value for whether this edge exists in the Complete Knowledge Graph i.e. we record whether there was a non-zero dropout percentage for a clinical trial due to an adverse event. Standard classifiers (Logistic Regression, Random Forest, Support Vector) were then trained on the array-structured data to predict edge indicator values using an 80\%-20\% train-test split, and ROC-AUC scores for the classifiers' predictions were generated and are provided in the Results section.

\subsection{General Implementation Details}

TrialGraph's methods and algorithms were implemented using programs which were written in Python 3 and run on a Jupyter Notebook environment with 2 GPU cores and 32 GB storage. Initial clinical trial data in CSV format was processed using Python's pandas library. MetaPath2Vec and HinSAGE methods use StellarGraph [11], a library providing graph-based machine learning tools, as a top level module incorporating several other machine learning libraries such as scikit-learn and TensorFlow. Clinical trial data was directly loaded into a StellarGraph graph format from pandas dataframes, and the MetaPath2Vec and HinSAGE algorithms were applied to produce predictions on test data using a combination of StellarGraph and scikit-learn/TensorFlow routines. Graph Kernel methods instead use GraKel [14] to load clinical trial data into a Grakel graph format from pandas dataframes, for which graph kernel functions were computed. Sci-kit learn was then used to train classifiers on graph feature representations to produce predictions. Hyperparameter training was performed for all methods\footnote{See Supplemental Data - (2)}. A full workflow diagram for TrialGraph is provided in Figure 7.

\begin{figure}
  \centering
  \includegraphics[width=\textwidth]{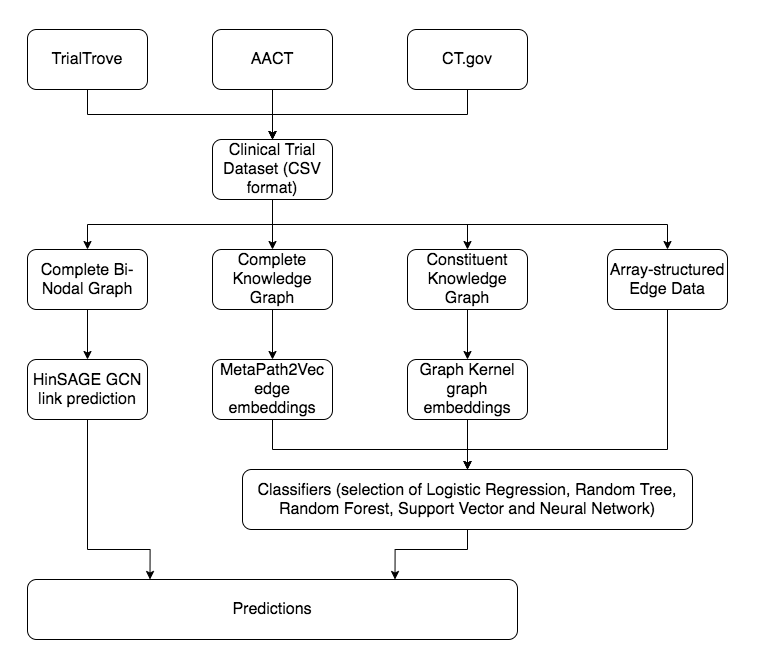}
  \caption{TrialGraph workflow diagram.}
\end{figure}

\section{Results}

All methods described in TrialGraph perform binary link classification to predict whether patients in a clinical trial are likely to suffer from a side effect. We can therefore analyze and compare the performance of these methods through Receiver Operating Characteristic (ROC) curves. These curves were plotted for the predictions of each method across several runs, and the area under each curve was calculated to produce ROC-AUC scores, which typically lie in the interval $[0.5,1]$. A score near $0.5$ is expected for a classifier which makes predictions randomly, and a score near $1$ indicates the classifier performs excellently. ROC-AUC scores for the methods described in TrialGraph are given in Tables \ref{table4} and \ref{table5}, and example ROC curves are provided in Figure 8. A kernel density estimation plot for the ROC-AUC scores of Graph Kernel methods across many adverse event predictions\footnote{See Supplemental Data - (3)} is given in Figure 9.

\begin{table}
  \caption{TrialGraph ROC-AUC scores}
  \label{table4}
  \centering
  \begin{tabular}{lccccc}
    \toprule
    Algorithm and Classifier & Run 1 & Run 2 & Run 3 & Mean & S.D.\\
    \midrule
    Array-Structured, Logistic Regression & 0.709 & 0.723 & 0.710 & 0.714 & 0.008\\
    Array-Structured, Random Forest & 0.705 & 0.706 & 0.694 & 0.702 & 0.007\\
    Array-Structured, Support Vector & 0.692 & 0.697 & 0.710 & 0.700 & 0.009\\
    MetaPath2Vec, Logistic Regression & 0.857 & 0.857 & 0.848 & 0.854 & 0.005\\
    MetaPath2Vec, Random Tree & 0.685 & 0.674 & 0.688 & 0.682 & 0.007\\
    MetaPath2Vec, Random Forest & 0.864 & 0.843 & 0.860 & 0.856 & 0.011\\
    MetaPath2Vec, Support Vector & 0.806 & 0.796 & 0.789 & 0.797 & 0.008\\
    MetaPath2Vec, Neural Network & 0.768 & 0.775 & 0.777 & 0.773 & 0.004\\
    HinSAGE & 0.796 & 0.800 & 0.797 & 0.798 & 0.002\\
    \bottomrule
  \end{tabular}
\end{table}

\begin{table}
\begin{minipage}{\textwidth}
  \caption{TrialGraph Graph Kernel ROC-AUC scores}
  \label{table5}
  \centering
  \begin{tabular}{lccccccc}
    \toprule
    Algorithm and Classifier & Run 1\textsuperscript{*} & Run 2\textsuperscript{*} & Run 3\textsuperscript{*} & Run 4\textsuperscript{*} & Run 5\textsuperscript{*} & Mean & S.D.\\
    \midrule
    Graph Kernel (ODD-STh) & 0.683 & 0.500 & 0.683 & 0.709 & 0.666 & 0.648 & 0.084\\
    Graph Kernel (PyramidMatch) & 0.757 & 0.724 & 0.654 & 0.705 & 0.662 & 0.700 & 0.043\\
    Graph Kernel (Propagation) & 0.701 & 0.543 & 0.700 & 0.691 & 0.634 & 0.654 & 0.068\\
    \bottomrule
    
  \end{tabular}
  \footnotetext{\textsuperscript{*}Adverse Events for Runs 1 through 5 were AE\_AST\_ALT, AE\_renal\_failure, AE\_thrombosis, AE\_sepsis}
  \footnotetext{and AE\_embolism respectively}
\end{minipage}
\end{table}

\begin{figure}
  \centering
  \includegraphics[width=\textwidth]{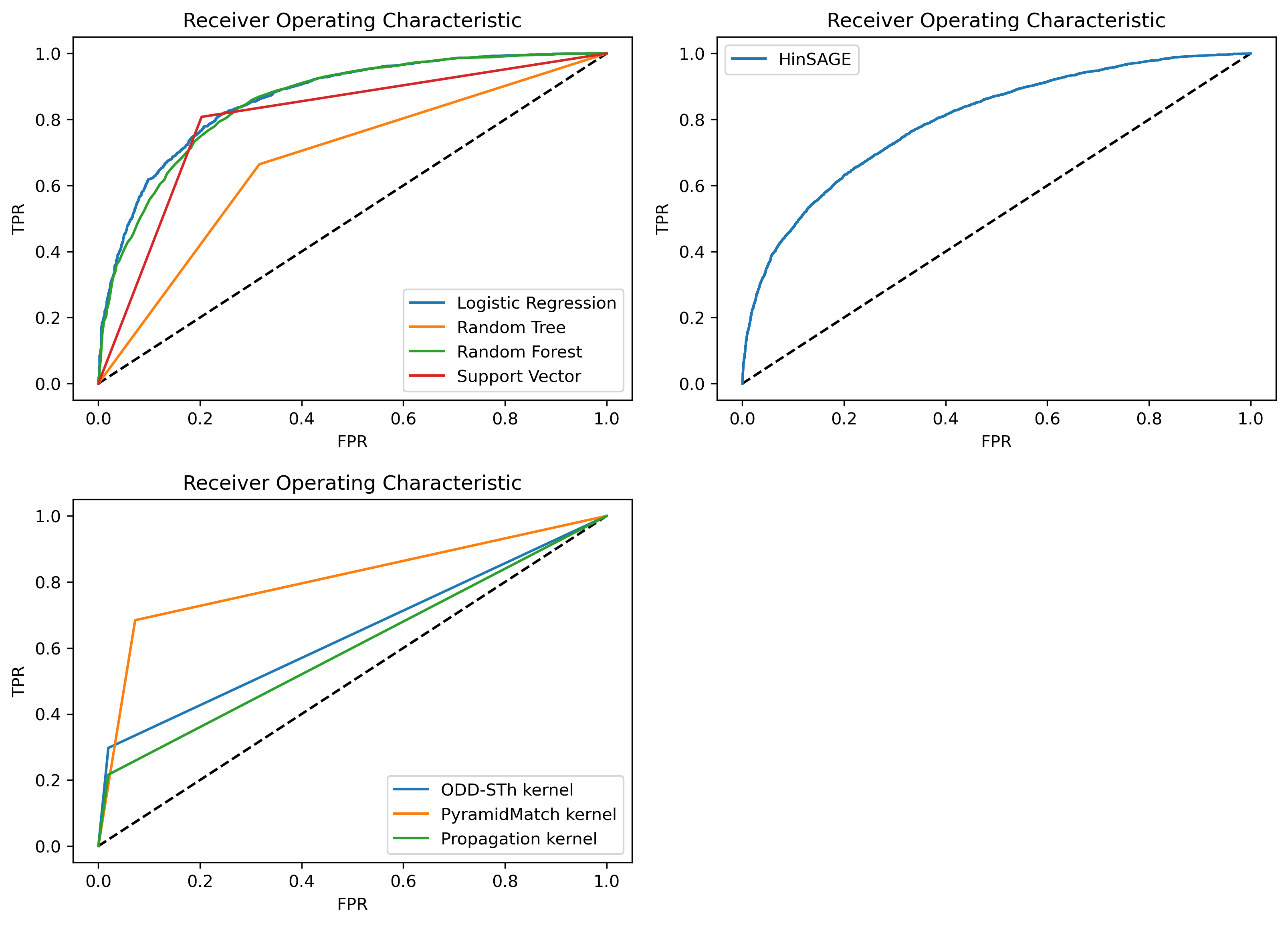}
  \caption{Example ROC curves for TrialGraph algorithms.}
\end{figure}

\begin{figure}
  \centering
  \includegraphics[width=0.75\textwidth]{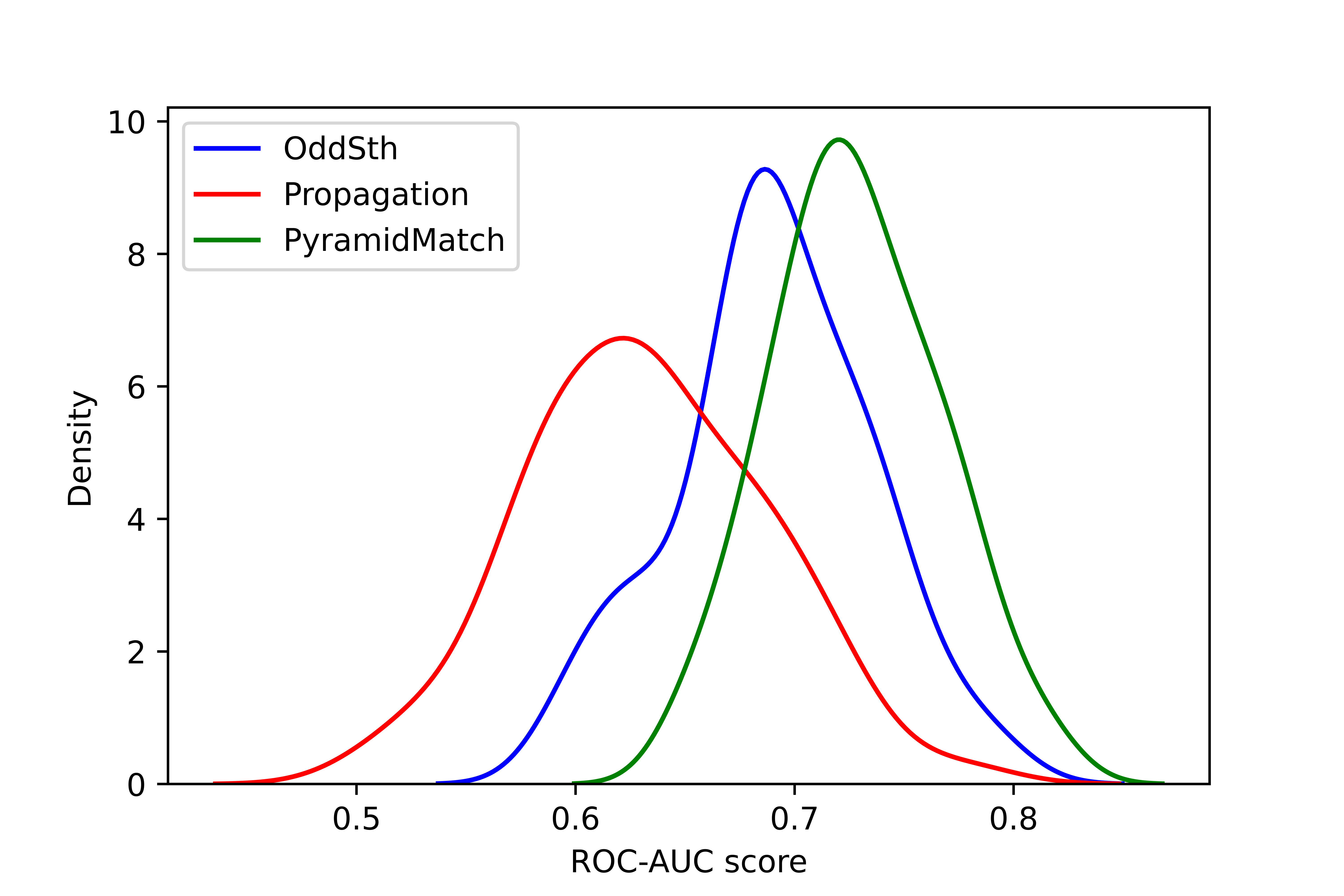}
  \caption{Kernel density estimation plots for sets of ROC-AUC scores generated for the prediction of each adverse event using each Graph Kernel method.}
\end{figure}

\section{Conclusion}

From the results provided in Tables 4 and 5, we conclude that graph-based machine learning algorithms perform well at edge classification tasks for clinical trial data, with most algorithms outperforming traditional array-structured approaches. The MetaPath2Vec algorithm with Logistic Regression and Random Forest classifiers produced the highest mean ROC-AUC scores of 0.85 and 0.86 respectively, with little score deviation across multiple runs indicating these methods produce accurate, reliable predictions. However, Metapath2Vec methods are limited by their transductive nature; in order to make edge classification predictions on new clinical trials, the entire model needs to be retrained on a new graph incorporating new data. The HinSAGE algorithm also performs well, with a mean ROC-AUC score of 0.80 significantly outperforming that of traditional array-structured approaches. While HinSAGE performs slightly worse at prediction tasks than MetaPath2Vec, it is inductive in nature which provides a significant advantage. Since aggregator functions are trained to create node representations based on the node's neighbourhood, embeddings for new clinical trial nodes can be quickly determined from links to the existing graph data. However, the HinSAGE algorithm only discriminates between node labels when training its aggregator functions, which makes it unsuitable for use with edge labelled graphs. Graph Kernel methods perform the weakest out of the graph-based machine learning techniques, and also occasionally produce very poor predictors e.g. ODD-STh kernel predicting AE\_renal\_failure, though this may be due to the specific nature of the prediction task. As indicated in Figure 9, the type of kernel function used significantly affects model performance; while the Propagation kernel performs poorly, the PyramidMatch kernel reliably produces better results than the array-structured comparison. Graph Kernel methods are inductive since representations for new clinical trial data can easily be determined by computing kernel functions with respect to a reference set of graphs. They are also the most generally applicable methods as they can operate on operate any graph format with the appropriate choice of kernel function, and are the quickest and simplest to run out of the methods TrialGraph uses.

\subsection{Discussion}

Graph representation of biomedical data is not a novel concept in biology [15]. Concepts, methods and tools from graph theory and network science have been applied to solve several challenging problems in biology, such as biological functional annotation, drug discovery, target discovery and drug repositioning [16-20]. While biological data has been extensively modelled using graph approaches [21, 22], its application in healthcare and clinical research has been limited. While approaches such as an EHR-oriented graph system and a COVID-19 trial graph have been proposed, most of the reported applications were limited to information extraction tasks [23-30]. To the best of our knowledge, TrialGraph is the first project of its kind to model clinical trial information that includes meta-data and results in a graph representation and use such knowledge graphs to mine massive-scale data. The premise of our work is based on the hypothesis that modelling clinical trial data as a graph and applying graph-based machine learning techniques can boost the performance of standard classifiers. Our initial results from the experimentation discussed in this paper suggest that such approaches could be useful. However, further research is required to increase graph complexity and augment clinical trial data with other information types including patient level information [31]. Currently, clinical trials are on the verge of a digital revolution enabled by big data, scalable computing and machine intelligence [32-36]; approaches such as TrialGraph could further enhance these initiatives by enabling graph-based modelling and inferences from historical clinical trial data for a therapeutic area or a disease of interest.

\subsection{Applications}

All TrialGraph models produce predictions on whether patients in a particular clinical trial will suffer side effects. This kind of prediction task has direct applications in industry; for example, such models could be used to predict which pre-clinical drugs are most likely to fail clinical trials. Pharmaceutical companies can then focus their efforts on the drugs that are most likely to succeed, avoiding the cost and delay of a failed clinical trial. However, TrialGraph primarily aims to provide a proof-of-concept framework for applying graph-based machine learning algorithms, and the models produced in TrialGraph would need to be significantly refined before they see actual industrial application (see Future Work). Despite this, TrialGraph is significant because many medical problems can be rephrased into some sort of graph prediction task, to which TrialGraph's methodology, if not its models, can be applied.

\subsection{Limitations}

The machine learning models produced in TrialGraph have a couple of limitations, the most significant of which is a lack of hyperparameter training. While hyperparameter training is very effective at improving the predictive power of machine learning algorithms, it is infeasible to perform in the context of TrialGraph since algorithms such as MetaPath2Vec generate embedding functions using several hyperparameters, and then apply several classifiers each with their own set of hyperparameters. Very significant computing power is required to train all hyperparameters for every combination of algorithm and classifier; if instead only a selection of hyperparameters are trained, it becomes harder to compare performances across algorithms and classifiers. Moreover, since several such algorithms produced good ROC-AUC scores, it is unnecessary to perform hyperparameter training to conclude that graph-based machine learning algorithms produce good results on clinical trial data, as hyperparameter training will only improve these ROC-AUC scores.

Restrictions in available computing power did not significantly affect the implementation of TrialGraph methods, with one minor exception. For certain choices of kernel function used in the graph kernel algorithm, the computational cost is $O(n^{6})$ or worse. These kernel functions are not used in TrialGraph as they are infeasible to implement.

\subsection{Future Work}

The methodology of extracting keywords as phrases which appear within multiple data entries to facilitate the construction of the Complete Knowledge Graph is sufficient for the purposes of TrialGraph, which is intended as a self-contained project on a curated clinical trial dataset. However, this approach neither generalises well to arbitrarily large datasets since some keywords need to be curated manually, nor leverages the wealth of medical knowledge available. Future versions of TrialGraph would aim to use external medical databases to enrich the graph structure by replacing keyword connections by biological pathways and encoding supplemental medical information as node attributes [37]. Future versions of TrialGraph might also perform different prediction tasks on different clinical trial data, especially if the scope of the project moves beyond exploratory to industry-focused. It is hoped that TrialGraph’s work on graph-based machine learning algorithms proves useful in this case and any other for which graph-structured data is relevant, and that these algorithms will go on to streamline clinical trials and ultimately benefit patients.

\section*{References}

\small

[1] DiMasi, J. A., Grabowski, H. G., \& Hansen, R. W. (2016). Innovation in the pharmaceutical industry: New estimates of R\&D costs. Journal of health economics, 47, 20–33. https://doi.org/10.1016/j.jhealeco.2016.01.012

[2] Shameer, K., Badgeley, M. A., Miotto, R., Glicksberg, B. S., Morgan, J. W., \& Dudley, J. T. (2017). Translational bioinformatics in the era of real-time biomedical, health care and wellness data streams. Briefings in bioinformatics, 18(1), 105–124. https://doi.org/10.1093/bib/bbv118

[3] Malik, L., \& Lu, D. (2019). Increasing complexity in oncology phase I clinical trials. Investigational new drugs, 37(3), 519–523. https://doi.org/10.1007/s10637-018-0699-1

[4] Shameer, K., Tripathi, L. P., Kalari, K. R., Dudley, J. T., \& Sowdhamini, R. (2016). Interpreting functional effects of coding variants: challenges in proteome-scale prediction, annotation and assessment. Briefings in bioinformatics, 17(5), 841–862. https://doi.org/10.1093/bib/bbv084

[5] Glicksberg, B. S., Li, L., Cheng, W. Y., Shameer, K., Hakenberg, J., Castellanos, R., Ma, M., Shi, L., Shah, H., Dudley, J. T., \& Chen, R. (2015). An integrative pipeline for multi-modal discovery of disease relationships. Pacific Symposium on Biocomputing. Pacific Symposium on Biocomputing, 20, 407–418.

[6] Peters, L. A., Perrigoue, J., Mortha, A., Iuga, A., Song, W. M., Neiman, E. M., Llewellyn, S. R., Di Narzo, A., Kidd, B. A., Telesco, S. E., Zhao, Y., Stojmirovic, A., Sendecki, J., Shameer, K., Miotto, R., Losic, B., Shah, H., Lee, E., Wang, M., Faith, J. J., … Schadt, E. E. (2017). A functional genomics predictive network model identifies regulators of inflammatory bowel disease. Nature genetics, 49(10), 1437–1449. https://doi.org/10.1038/ng.3947

[7] Johnson, K. W., Shameer, K., Glicksberg, B. S., Readhead, B., Sengupta, P. P., Björkegren, J., Kovacic, J. C., \& Dudley, J. T. (2017). Enabling Precision Cardiology Through Multiscale Biology and Systems Medicine. JACC. Basic to translational science, 2(3), 311–327. https://doi.org/10.1016/j.jacbts.2016.11.010

[8] Xiang Yue, Zhen Wang, Jingong Huang, Srinivasan Parthasarathy, Soheil Moosavinasab, Yungui Huang, Simon M Lin, Wen Zhang, Ping Zhang, Huan Sun, Graph embedding on biomedical networks: methods, applications and evaluations, Bioinformatics, Volume 36, Issue 4, 15 February 2020, Pages 1241–1251, https://doi.org/10.1093/bioinformatics/btz718

[9] Thomas Gaudelet, Ben Day, Arian R Jamasb, Jyothish Soman, Cristian Regep, Gertrude Liu, Jeremy B R Hayter, Richard Vickers, Charles Roberts, Jian Tang, David Roblin, Tom L Blundell, Michael M Bronstein, Jake P Taylor-King, Utilizing graph machine learning within drug discovery and development, Briefings in Bioinformatics, 2021;, bbab159, https://doi.org/10.1093/bib/bbab159

[10] Yuxiao Dong, Nitesh V. Chawla, and Ananthram Swami. 2017. Metapath2vec: Scalable Representation Learning for Heterogeneous Networks. In Proceedings of the 23rd ACM SIGKDD International Conference on Knowledge Discovery and Data Mining (KDD '17). Association for Computing Machinery, New York, NY, USA, 135–144. DOI:https://doi.org/10.1145/3097983.3098036

[11] CSIRO's Data61. 2018. StellarGraph Machine Learning Library. GitHub, GitHub Repository. https://github.com/stellargraph/stellargraph

[12] William L. Hamilton and Rex Ying and Jure Leskovec. 2018. Inductive Representation Learning on Large Graphs. https://arxiv.org/abs/1706.02216

[13] Karsten Borgwardt, Elisabetta Ghisu, Felipe Llinares-López, Leslie O’Bray and Bastian Rieck (2020), "Graph Kernels: State-of-the-Art and Future Challenges", Foundations and Trends® in Machine Learning: Vol. 13: No. 5-6, pp 531-712. http://dx.doi.org/10.1561/2200000076

[14] Giannis Siglidis, Giannis Nikolentzos, Stratis Limnios, Christos Giatsidis, Konstantinos Skianis, Michalis Vazirgiannis. 2020. GraKeL: A Graph Kernel Library in Python. https://arxiv.org/abs/1806.02193

[15] A. L. Barabasi, Z. N. Oltvai, "Network biology: understanding the cell's functional organization," Nat Rev Genet, vol. 5, no. 2, pp. 101-13, Feb 2004, doi: 10.1038/nrg1272.

[16] P. Barah, N. M. B, N. D. Jayavelu, R. Sowdhamini, K. Shameer, and A. M. Bones, "Transcriptional regulatory networks in Arabidopsis thaliana during single and combined stresses," Nucleic Acids Res, vol. 44, no. 7, pp. 3147-64, Apr 20 2016, doi: 10.1093/nar/gkv1463.

[17] K. Shameer et al., "A Network-Biology Informed Computational Drug Repositioning Strategy to Target Disease Risk Trajectories and Comorbidities of Peripheral Artery Disease," AMIA Jt Summits Transl Sci Proc, vol. 2017, pp. 108-117, 2018. [Online].

[18] L. A. Peters et al., "A functional genomics predictive network model identifies regulators of inflammatory bowel disease," Nat Genet, vol. 49, no. 10, pp. 1437-1449, Oct 2017, doi: 10.1038/ng.3947.

[19] B. S. Glicksberg et al., "Comparative analyses of population-scale phenomic data in electronic medical records reveal race-specific disease networks," Bioinformatics, vol. 32, no. 12, pp. i101-i110, Jun 15 2016, doi: 10.1093/bioinformatics/btw282.

[20] K. Shameer, M. Naika, O. K. Mathew, and R. Sowdhamini, "Network Modules Driving Plant Stress Response, Tolerance and Adaptation: A case study using Abscisic acid Induced Protein-protein Interactome of <em>Arabidopsis thaliana</em>," bioRxiv, p. 073247, 2016, doi: 10.1101/073247.

[21] P. Shannon et al., "Cytoscape: a software environment for integrated models of biomolecular interaction networks," Genome Res, vol. 13, no. 11, pp. 2498-504, Nov 2003, doi: 10.1101/gr.1239303.

[22] D. Otasek, J. H. Morris, J. Boucas, A. R. Pico, and B. Demchak, "Cytoscape Automation: empowering workflow-based network analysis," Genome Biol, vol. 20, no. 1, p. 185, Sep 2 2019, doi: 10.1186/s13059-019-1758-4.

[23] J. R. Petrella, "Use of graph theory to evaluate brain networks: a clinical tool for a small world?," Radiology, vol. 259, no. 2, pp. 317-20, May 2011, doi: 10.1148/radiol.11110380.

[24] S. G. Finlayson, P. LePendu, and N. H. Shah, "Building the graph of medicine from millions of clinical narratives," Sci Data, vol. 1, p. 140032, 2014, doi: 10.1038/sdata.2014.32.

[25] Z. Haneef and S. Chiang, "Clinical correlates of graph theory findings in temporal lobe epilepsy," Seizure, vol. 23, no. 10, pp. 809-18, Nov 2014, doi: 10.1016/j.seizure.2014.07.004.

[26] D. Kim et al., "Knowledge boosting: a graph-based integration approach with multi-omics data and genomic knowledge for cancer clinical outcome prediction," J Am Med Inform Assoc, vol. 22, no. 1, pp. 109-20, Jan 2015, doi: 10.1136/amiajnl-2013-002481.

[27] D. Birtwell, H. Williams, R. Pyeritz, S. Damrauer, and D. L. Mowery, "Carnival: A Graph-Based Data Integration and Query Tool to Support Patient Cohort Generation for Clinical Research," Stud Health Technol Inform, vol. 264, pp. 35-39, Aug 21 2019, doi: 10.3233/SHTI190178.

[28] J. Du et al., "COVID-19 trial graph: a linked graph for COVID-19 clinical trials," J Am Med Inform Assoc, vol. 28, no. 9, pp. 1964-1969, Aug 13 2021, doi: 10.1093/jamia/ocab078.

[29] Y. Shang et al., "EHR-Oriented Knowledge Graph System: Toward Efficient Utilization of Non-Used Information Buried in Routine Clinical Practice," IEEE J Biomed Health Inform, vol. 25, no. 7, pp. 2463-2475, Jul 2021, doi: 10.1109/JBHI.2021.3085003.

[30] K. Zhan et al., "Novel Graph-Based Model With Biaffine Attention for Family History Extraction From Clinical Text: Modeling Study," JMIR Med Inform, vol. 9, no. 4, p. e23587, Apr 21 2021, doi: 10.2196/23587.

[31] Y. Zhang, S. Nampally, E. Hutchison, J. Weatherall, F. Khan, and K. Shameer, "Predictive Modeling of Personalized Clinical Trial Attrition using Time-to-event Approaches," in Machine Learning for Pharma and Healthcare Applications ECML PKDD 2020 Workshop; PharML, 2020, vol. 2020.

[32] E. H. Weissler et al., "The role of machine learning in clinical research: transforming the future of evidence generation," Trials, vol. 22, no. 1, p. 537, Aug 16 2021, doi: 10.1186/s13063-021-05489-x.

[33] H. Lea et al., "Machine Learning Can Enable the Automation of Clinical Endpoint Adjudication," Circulation, vol. 144, no. Suppl\_1, pp. A9330-A9330, 2021, doi: doi:10.1161/circ.144.suppl\_1.9330.

[34] H. Lea et al., "Can machine learning augment clinician adjudication of events in cardiovascular trials? A case study of major adverse cardiovascular events (MACE) across CVRM trials," European Heart Journal, vol. 42, no. Supplement\_1, 2021, doi: 10.1093/eurheartj/ehab724.3061.

[35] A. Prokop, Y. Zhang, P. Mukhopadhyay, F. Khan, and K. Shameer, "SAEgnal: A Predictive Assessment Framework for Optimizing Safety Profiles in Immuno-Oncology Combination Trials," AMIA Jt Summits Transl Sci Proc, vol. 2021, pp. 535-544, 2021. [Online]

[36] E. R. Hutchison, Y. Zhang, S. Nampally, J. Weatherall, F. Khan, and K. Shameer, "Uncovering Machine Learning-Ready Data from Public Clinical Trial Resources: A case-study on normalization across Aggregate Content of ClinicalTrials.gov," in 2020 IEEE International Conference on Bioinformatics and Biomedicine (BIBM), 16-19 Dec. 2020 2020, pp. 2965-2967, doi: 10.1109/BIBM49941.2020.9313362.

[37] Stephen Bonner, Ian P Barrett, Cheng Ye, Rowan Swiers, Ola Engkvist, Andreas Bender, Charles Tapley Hoyt, William Hamilton. 2021. A Review of Biomedical Datasets Relating to Drug Discovery: A Knowledge Graph Perspective. https://arxiv.org/abs/2102.10062

\newpage

\section*{Supplemental Data}

(1) - Metapaths used in MetaPath2Vec algorithm are given as follows:

['Clinical Trial','Side Effect','Clinical Trial'],
['Clinical Trial','Specific Drug','Clinical Trial'],
['Clinical Trial','Drug','Specific Drug','Clinical Trial'],
['Clinical Trial','Specific Drug','Drug','Clinical Trial'],
['Clinical Trial','Specific Disease','Clinical Trial'],
['Clinical Trial','Disease','Specific Disease','Clinical Trial'],
['Clinical Trial','Specific Disease','Disease','Clinical Trial'],
['Drug','Specific Drug','Drug'],
['Specific Drug','Drug','Specific Drug'],
['Disease','Specific Disease','Disease'],
['Specific Disease','Disease','Specific Disease'],
['Drug','Disease','Drug'],
['Disease','Drug','Disease'],
['Specific Drug','Disease','Specific Drug'],
['Specific Disease','Drug','Specific Disease'],
['Side Effect','Clinical Trial','Side Effect']

(2) - Hyperparameters were trained using grid search and a standard scaler. The hyperparameters trained for each of the different classifiers and algorithms are given in Table 6:

(3) - ROC-AUC scores were generated when each of the kernel functions OddSth, PyramidMatch and Propagation were used in the Graph Kernel method to produce predictions for each Adverse Event, and are given in Table 7:

\newpage

\begin{table}[H]
  \caption{Hyperparameter training variables}
  \label{table6}
  \centering
  \begin{tabularx}{\textwidth}{|X|X|}
    \toprule
    Classifier & Variables \\
    \midrule
    Logistic Regression CV & 'max\_iter'\\
    Random Tree & 'max\_depth'\\
    Random Forest & 'max\_depth', 'n\_estimators'\\
    Support Vector & 'C', 'gamma', 'kernel'\\
    Neural Network & -\\
    \midrule
    Algorithm & Variables\\
    \midrule
    MetaPath2Vec & 'Binary Operator'\\
    HinSAGE & -\\
    Graph Kernel & -\\
    \bottomrule
  \end{tabularx}
\end{table}

\pagenumbering{gobble}

\begin{table}[H]
\caption{ROC-AUC scores for Graph Kernel predictions of Adverse Event prevalence}
\label{table}
\small
\centering
\begin{tabular}{l|c|c|c}%
\bfseries Predicted Adverse Event & \bfseries OddSth Kernel & \bfseries PyramidMatch Kernel & \bfseries Propagation Kernel
\csvreader[head to column names]{transposed_full_data.csv}{}
{\\\hline\AdverseEvent & \OddSth & \PyramidMatch & \Propagation}
\end{tabular}
\end{table}

\end{document}